\documentclass[10pt,twocolumn,letterpaper]{article}

\usepackage{cvpr}              % To produce the CAMERA-READY version
\usepackage{booktabs}
\usepackage{multirow}
\usepackage{colortbl}
\usepackage{arydshln}
\usepackage{amsmath}
\usepackage{pifont}
\usepackage[most]{tcolorbox}

\definecolor{yellowhighlight}{RGB}{255,255,204}
\definecolor{bluehighlight}{RGB}{221,235,247}
\definecolor{lightgray}{gray}{0.9}
\definecolor{graytext}{gray}{0.5}
\definecolor{greenhighlight}{RGB}{210,235,200}
\newcommand{\cmark}{\textcolor{red}{\ding{51}}}      % 红色勾
\newcommand{\xmark}{\textcolor{green!60!black}{\ding{55}}} % 绿色叉
\usepackage{soul}
\newcommand{\blue}[1]{\textcolor{blue}{#1}}

\newcommand\blfootnote[1]{%
  \begingroup
  \renewcommand\thefootnote{}\footnote{#1}%
  \addtocounter{footnote}{-1}%
  \endgroup
}

\definecolor{cvprblue}{rgb}{0.21,0.49,0.74}
\usepackage[pagebackref,breaklinks,colorlinks,allcolors=cvprblue]{hyperref}

\def\paperID{16611} % *** Enter the Paper ID here
\def\confName{CVPR}
\def\confYear{2026}

\title{DiG: Differential Grounding for Enhancing Fine-Grained Perception in Multimodal Large Language Models}

\author{
Zhou Tao\textsuperscript{1,2$*$},
Shida Wang\textsuperscript{1,2$*$}, 
Yongxiang Hua\textsuperscript{1,2},
Haoyu Cao\textsuperscript{1,2}, 
Linli Xu\textsuperscript{1,2$\dagger$} \\
\textsuperscript{1}University of Science and Technology of China\\
\textsuperscript{2}State Key Laboratory of Cognitive Intelligence\\
{\tt\small{
\{zhoutao24, wangshida, yx15333063290, caohaoyu\}@mail.ustc.edu.cn
}}\\
{\tt\small{linlixu@ustc.edu.cn}}
}

\begin{document}
\maketitle
\blfootnote{$^*$ Equal contribution.\ \ $\dagger$ Corresponding author. }

\begin{abstract}
%Multimodal Large Language Models have achieved impressive performance on a variety of vision-language tasks, yet their fine-grained visual perception and precise spatial reasoning remain limited. In this work, we introduce DiG (Differential Grounding), a novel proxy task framework designed to enhance fine-grained perceptual capabilities in MLLMs. DiG presents models with pairs of highly similar images and requires the identification and localization of all visual discrepancies without prior knowledge of their number. To support scalable training, we develop an automated 3D rendering-based data generation pipeline that produces high-quality paired images with fully controllable discrepancies. We further employ a Curriculum Learning strategy to overcome the sparsity of difference signals and enable stable optimization. Extensive experiments demonstrate that DiG significantly improves model performance across a variety of visual perception benchmarks and that the learned fine-grained perception skills transfer effectively to standard downstream tasks, including RefCOCO, RefCOCO+, RefCOCOg, and general multimodal perception benchmarks. Our results highlight differential grounding as a scalable and robust approach for advancing fine-grained visual reasoning in MLLMs.

Multimodal Large Language Models have achieved impressive performance on a variety of vision-language tasks, yet their fine-grained visual perception and precise spatial reasoning remain limited. In this work, we introduce DiG (Differential Grounding), a novel proxy task framework where MLLMs learn fine-grained perception by identifying and localizing all differences between similar image pairs without prior knowledge of their number. To support scalable training, we develop an automated 3D rendering-based data generation pipeline that produces high-quality paired images with fully controllable discrepancies. To address the sparsity of difference signals, we further employ curriculum learning that progressively increases complexity from single to multiple differences, enabling stable optimization. Extensive experiments demonstrate that DiG significantly improves model performance across a variety of visual perception benchmarks and that the learned fine-grained perception skills transfer effectively to standard downstream tasks, including RefCOCO, RefCOCO+, RefCOCOg, and general multimodal perception benchmarks. Our results highlight differential grounding as a scalable and robust approach for advancing fine-grained visual reasoning in MLLMs.

\end{abstract}
    
\section{Introduction}
\label{sec:intro}

\begin{figure}[t]
  \centering
   \includegraphics[width=\linewidth]{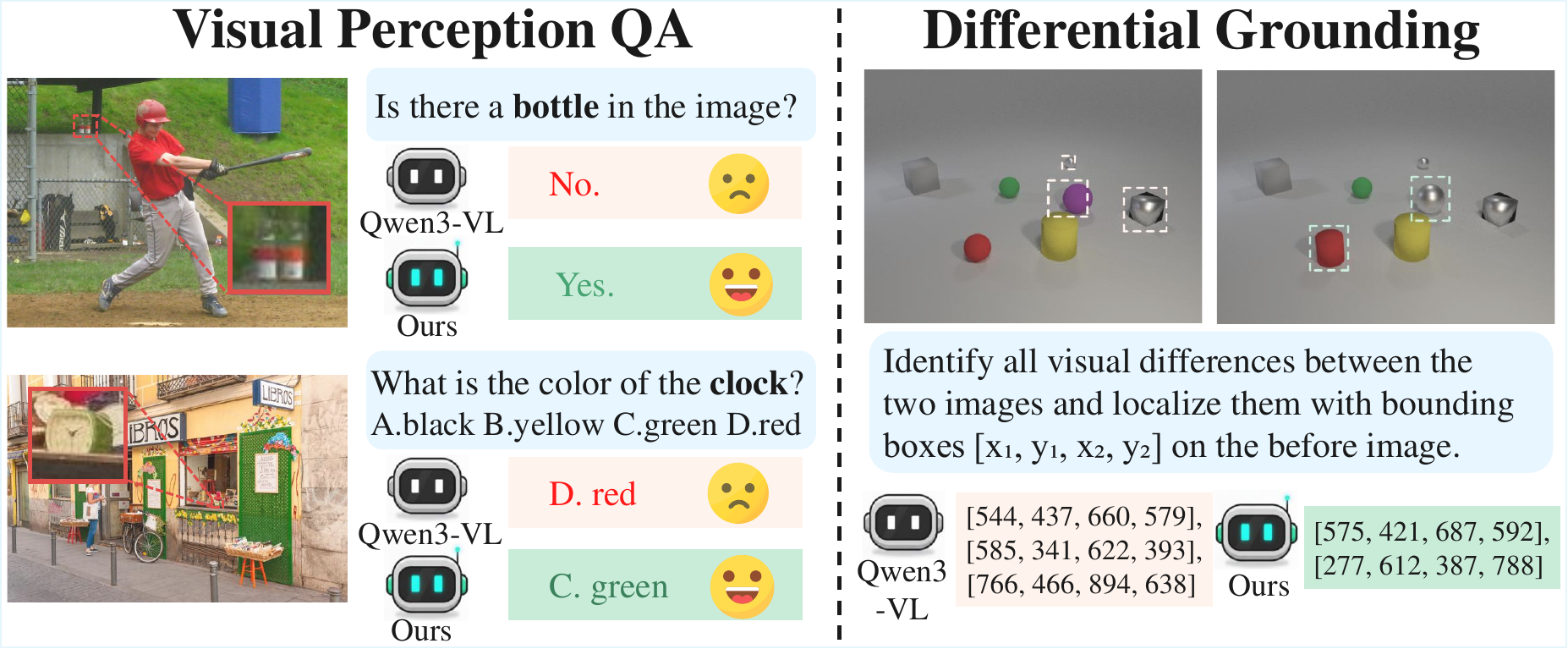}
   \caption{Illustration of Differential Grounding (DiG). \blue{Left}: existing MLLMs fail to perceive subtle visual cues such as small object changes or color variations. \blue{Right}: our proposed DiG task trains models to identify and localize differences between paired images, fostering fine-grained visual perception.}
   \label{fig:intro}
\end{figure}

Multimodal Large Language Models (MLLMs) have achieved remarkable progress across diverse vision-language tasks, including image captioning and visual question answering~\cite{bai2025qwen25vltechnicalreport,wang2025internvl3,chen2024sharegpt4v,hurst2024gpt,comanici2025gemini,bai2023qwenvlversatilevisionlanguagemodel,liu2024hrvda,cao2023attention,hua2025input}. While these
%These 
models excel at holistic scene understanding, effectively associating global visual semantics with textual reasoning, they %. However, as illustrated in the upper part of \cref{fig:intro}, they still
struggle with fine-grained visual perception and precise spatial reasoning. As illustrated in Figure~\ref{fig:intro} (left), even %Even 
state-of-the-art MLLMs often overlook subtle visual cues such as small object changes, color variations, or missing elements that are essential for detailed visual understanding. 
This highlights a fundamental limitation of current pretraining paradigms: while they enable strong high-level semantic alignment, they provide insufficient supervision for developing the fine-grained perceptual sensitivity required for nuanced scene analysis.

Reinforcement Learning (RL)~\cite{schulman2017proximal,rafailov2023direct} has recently emerged as a powerful paradigm for enhancing the multimodal capabilities of large models.
Beyond conventional supervised training, RL-based post-training has proven effective in improving robustness, alignment, and perceptual fidelity, and is now widely adopted for refining MLLMs~\cite{zhang2025r1vllearningreasonmultimodal,tan2025reason,meng2025mm,huang2025vision}.
Building on this progress, recent studies have explored leveraging RL to strengthen visual perception in multimodal models.
However, existing approaches often involve trade-offs between task specificity and generalization, or rely on costly, manually annotated datasets that limit scalability and diversity.
One line of work fine-tunes models directly on traditional perceptual tasks, such as visual grounding~\cite{yu2025perception}.
While this approach can boost task-specific accuracy, it often produces overspecialized models that generalize poorly to open-ended multimodal understanding.
Moreover, grounding datasets such as RefCOCO~\cite{yu2016modeling,kazemzadeh-etal-2014-referitgame} require extensive manual annotation, making large-scale training costly and limited in diversity. Another line of research explores alternative proxy tasks, such as visual jigsaw puzzles~\cite{wang2025jigsaw,wu2025visual,zeng2025agentic} or caption hallucination detection~\cite{wang2025vicrit}.
Although these tasks provide useful auxiliary supervision, they are not explicitly designed to strengthen fine-grained visual perception, leaving MLLMs insufficiently sensitive to subtle visual details.

To address these limitations, we propose \textbf{DiG} (Differential Grounding), a proxy framework designed to enhance fine-grained visual perception in MLLMs through a differential image grounding task.
Given two closely related images, the model is required to identify and localize all visual discrepancies without being informed of their exact number.
This formulation compels the model to move beyond coarse semantic alignment and develop fine-grained perceptual reasoning.
To %support this task
enable scalable training, we introduce a %scalable and
fully automated data generation pipeline based on 3D rendering~\cite{johnson2017clevr}.
%The
This pipeline allows precise control over the task difficulty by varying the number, subtlety, location, and type of visual differences, while ensuring accurate ground-truth annotations.
As a result, it enables the construction of large-scale, high-quality datasets that provide a strong foundation for improving the perceptual capabilities of MLLMs.

A key challenge in training MLLMs on the DiG task is the extreme sparsity of difference signals: at initialization, the model rarely localizes any differences correctly, causing RL rewards to collapse near zero and preventing stable optimization. %, which makes direct optimization difficult and causes reward functions in RL-based training to remain near zero at the start.
To address this, we employ a Curriculum Learning~\cite{bengio2009curriculum} (CL) strategy, progressively increasing task difficulty from simpler instances with a single discrepancy to more complex cases with multiple discrepancies.
This staged approach enables stable convergence by building perceptual skills from simple to complex cases, allowing models to develop fine-grained spatial reasoning without being overwhelmed by sparse supervision. %and allows the model to gradually acquire fine-grained visual reasoning skills in a scalable manner.
%By structuring the training in this way, the model can efficiently learn to handle tasks of varying complexity without being overwhelmed by sparse supervision.

We evaluate the effectiveness of DiG through extensive experiments.
Our approach consistently improves model performance on a range of visual perception benchmarks~\cite{guan2024hallusionbench,wang2025divide,li2023evaluating,wu2023vguidedvisualsearch,liu2023visual,tong2024cambrian,tong2024eyes}.
More importantly, the fine-grained visual perception skills acquired through DiG transfer effectively to standard downstream tasks, including fine-grained grounding benchmarks such as RefCOCO, RefCOCO+~\cite{yu2016modeling,kazemzadeh-etal-2014-referitgame}, and RefCOCOg~\cite{mao2016generation}, as well as general multimodal perception benchmarks~\cite{liu2024mmbench,yu2023mm,chen2024we,lu2022learn,singh2019towards,fu2023mme,kembhavi2016diagram}.
These results demonstrate the potential of differential grounding as a scalable and robust proxy task for developing fine-grained perceptual capabilities in MLLMs. Our main contributions are:

\begin{itemize}
    \item We introduce DiG (Differential Grounding), a novel proxy task framework that enhances fine-grained spatial perception in MLLMs, and develop a scalable, automated data generation pipeline that produces high-quality paired images with fully controllable discrepancies.
    \item We propose a Curriculum Learning strategy that addresses the optimization challenges posed by the sparsity of difference signals, enabling stable and effective training on the DiG task.
    \item We demonstrate through extensive experiments that models trained with DiG achieve significant improvements on a range of visual perception benchmarks and exhibit strong transferability to standard fine-grained grounding datasets as well as general multimodal perception benchmarks.
\end{itemize}

\begin{figure*}[t]
  \centering
   \includegraphics[width=\linewidth]{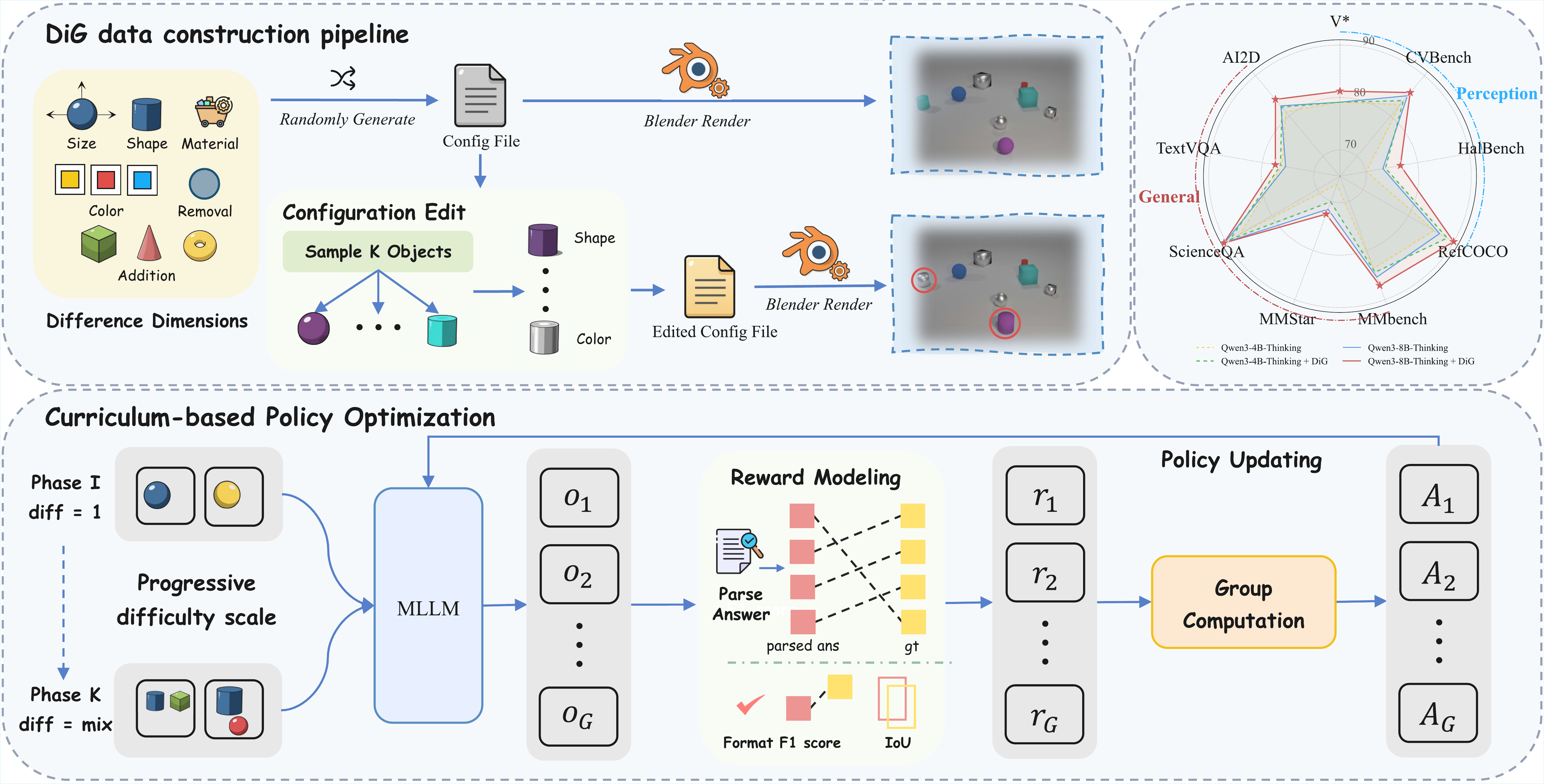}
   \caption{Overview of the proposed DiG framework.
The upper part illustrates the automated 3D-based data construction pipeline that generates paired images with controllable visual differences. The lower part shows the curriculum-based policy optimization, where the MLLM is trained via GRPO with rewards for format validity, detection accuracy, and spatial precision. The radar chart on the right demonstrates consistent performance gains across both perception and general multimodal benchmarks.}
   \label{fig:method}
\end{figure*}

\section{Related Work}
\label{sec:2_related_work}

\paragraph{Reinforcement Learning for MLLMs.}

Reinforcement learning (RL)~\cite{schulman2017proximal,rafailov2023direct,lambert2024tulu,chang2025thor} has emerged as a powerful paradigm for enhancing the capabilities of multimodal large language models.
With the success of rule-based RL~\cite{guo2025deepseek,shao2024deepseekmath,team2025kimi}, recent multimodal work explores RL with verifiable rewards, typically following a pipeline that conducts optional SFT activation then applies RL, such as GRPO~\cite{shao2024deepseekmath}, with fine-grained training recipes. 
Building on these advances, the MLLM community has begun extending this paradigm to enhance both visual perception and reasoning. This has spurred efforts in both dataset curation and method development.
On the data front, Vision-R1~\cite{huang2025vision} and MM-Eureka~\cite{meng2025mm} introduced large-scale SFT and RL datasets to facilitate effective cold-start training.
Methodologically, ReasonRFT~\cite{tan2025reason} and R1-VL~\cite{zhang2025r1vllearningreasonmultimodal} proposed novel rule-based reward functions to improve geometric understanding and spatial reasoning.
Furthering this direction, Perception-R1~\cite{yu2025perception} incorporated object-matching and IoU-based reward signals to enhance fine-grained grounding performance, marking a promising step toward perceptually-aligned reinforcement learning for MLLMs.

\paragraph{Generalization through Proxy-task-driven Learning.}
Proxy tasks have recently emerged as an effective strategy for improving generalization and reasoning in large models, offering controllable and verifiable supervision without costly human annotation. 
In language domains, several studies have demonstrated that solving structured synthetic problems, such as logic puzzles or rule-based games, through reinforcement learning can foster transferable reasoning abilities~\cite{xie2025logic,chen2025enigmata,dong2025reinforcement}. 
Motivated by these successes, recent multimodal research has explored analogous ideas using visually grounded proxies, including programmatically generated puzzles~\cite{wang2025jigsaw,wu2025visual,zeng2025agentic} and hallucination detection tasks~\cite{wang2025vicrit}. 
While these approaches provide valuable intermediate supervision, most remain limited to coarse or task-specific perception. 
Our work extends this proxy-learning perspective to fine-grained visual understanding, formulating difference localization as a scalable, automatically verifiable grounding problem.

\section{Method}

Our method comprises four components. We first formalize the Differential Grounding (DiG) task, then describe a 3D-based data pipeline for generating paired images with controlled discrepancies. We further introduce a reinforcement learning objective with structured rewards and a curriculum-based training strategy to ensure stable and effective optimization.

\subsection{Problem Formulation}
\label{sec:problem_formulation}

The core of our approach is the Differential Grounding (DiG) task, which trains MLLMs to identify and localize all visual differences between image pairs.
Given an input tuple X = ($I_a$, $I_b$, $P$) , where $I_a$ denotes a reference image and $I_b$ corresponds to its visually similar counterpart containing $M$ subtle modifications (e.g., additions, removals, or attribute changes in objects), the model receives a textual instruction $P$ that prompts it to identify and localize all discrepancies between the two images.

The model generates a textual output sequence $O = (o_1, o_2, \ldots, o_T)$, which can be deterministically parsed into a set of predicted bounding boxes $B_{\text{pred}} = \{b_i\}_{i=1}^{N}$, where each $b_i = [x_{\min}, y_{\min}, x_{\max}, y_{\max}]_i$ represents a localized discrepancy.
The corresponding ground-truth annotations are given by $B_{\text{gt}} = \{ b^{\text{gt}}_j \}_{j=1}^{M}$, which precisely mark all true differences between $I_a$ and $I_b$.

The objective of DiG is to train the model such that $B_{\text{pred}}$ aligns closely with $B_{\text{gt}}$ in both completeness and spatial precision.
Formally, this entails optimizing an objective function $\mathcal{L}_{DiG}$
 that encourages the model to perform fine-grained visual comparison, robust difference identification, and accurate spatial grounding within an open-ended generative framework.

\subsection{DiG Data Construction Pipeline}
\label{sec:data_construction}

To enable large-scale training for the DiG task, we construct a fully automated and scalable data generation pipeline based on the Blender 3D rendering engine.
This pipeline synthesizes paired images with precisely controlled visual discrepancies and automatically produces accurate ground-truth annotations, as illustrated in \cref{fig:method}.

The process begins with the procedural generation of a base 3D scene, where a configuration file in JSON format specifies the number of objects together with their geometric and visual attributes, including shape, material, color, size, and spatial position.
This configuration file is then rendered using Blender to produce the reference image $I_a$.

To generate the corresponding counterpart $I_b$, a stochastic modification process is applied to the base configuration.
An integer $K$ is randomly sampled from the range $[1, N]$, where $N$ denotes the maximum number of differences allowed in a scene.
A total of $K$ distinct objects are then selected, and for each object, one attribute is randomly sampled from a predefined set of modification types.
This set includes changes in shape, color, size, or material, as well as object addition or removal.
Applying these edits to the configuration file results in a modified version, which is subsequently rendered to obtain $I_b$.
The resulting image pair $(I_a, I_b)$ is visually coherent yet differs in exactly $K$ localized regions, ensuring both realism and controllable variability across the dataset.
In addition, the overall task difficulty can be flexibly extended by increasing the upper bound $N$, which allows the pipeline to generate more complex image pairs without altering the underlying process.

A major advantage of this rendering-based design is the ability to generate perfect ground-truth annotations automatically.
Because every modification is programmatically defined in three-dimensional space, its two-dimensional projection can be extracted directly from the renderer to form the bounding-box annotations $B_{\text{gt}}$.
This automatic supervision entirely eliminates the need for manual labeling and guarantees accurate and consistent annotations across all generated samples.

\subsection{Reward Modeling and Policy Optimization}
\label{sec:reward_policy}

We adopt a reinforcement learning paradigm to optimize the model on the DiG task using Group Relative Policy Optimization~\cite{shao2024deepseekmath} (GRPO). 
Given a prompt $P$ and image pair $(I_a, I_b)$, the model produces a textual output $O$, which is parsed into predicted bounding boxes $B_{\text{pred}}$.
The reward function consists of three complementary components that assess syntactic validity, detection accuracy, and spatial precision.

\noindent\textbf{Format Reward.}
Since the DiG task is formulated as an open-ended generation problem, ensuring syntactic correctness is a prerequisite for any subsequent evaluation. 
We define a \textit{format reward} $r_{\text{format}}$ to encourage the model to output responses that can be deterministically parsed into valid bounding-box lists. 
Let $\mathcal{P}(O)$ denote a parser that extracts $B_{\text{pred}}$ from the model’s textual output $O$.
The reward is a binary indicator that assigns a positive signal only when the output conforms to the target schema:
\begin{equation}
r_{\text{format}} =
\begin{cases}
1, & \text{if } \mathcal{P}(O)\neq \varnothing,\\
0, & \text{otherwise.}
\end{cases}
\label{eq:format_reward}
\end{equation}
This component serves as a structural constraint, guiding the model to produce well-formed and machine-interpretable responses such as 
``\texttt{[[x\textsubscript{min}, y\textsubscript{min}, x\textsubscript{max}, y\textsubscript{max}], ...]}'' rather than free-form text.

\noindent\textbf{Accuracy Reward.}
While the format reward enforces structural correctness, it does not reflect the perceptual quality of the predicted bounding boxes. 
To capture the model’s fine-grained localization and detection performance, we define an \textit{accuracy reward} $r_{\text{acc}}$ that jointly measures the completeness and precision of detected discrepancies. 
Given the set of predicted boxes $B_{\text{pred}}=\{b_i\}_{i=1}^{N}$ and ground-truth boxes $B_{\text{gt}}=\{\tilde b_j\}_{j=1}^{M}$, we first compute a one-to-one correspondence between predictions and ground truths to ensure reliable evaluation. 
This is formulated as a bipartite matching problem that maximizes the overall similarity between matched pairs, solved efficiently by the Hungarian algorithm~\cite{kuhn1955hungarian}. 
Each similarity score $\Phi(b_i,\tilde b_j)$ is defined as a combination of L1 distance and generalized intersection-over-union (GIoU) between the two boxes, and the matched pairs $\mathcal{M}$ are used to calculate both the detection and localization metrics.

The accuracy reward combines two complementary signals: a detection-level F1 score that reflects how completely the model identifies all differences, and an IoU term that captures the spatial alignment quality of matched boxes. 
Specifically, the F1 score is computed from precision $p=n_{\text{m}}/N$ and recall $r=n_{\text{m}}/M$, where $n_{\text{m}}$ denotes the number of matched pairs obtained by the Hungarian assignment. 
The F1 score is given by $\mathrm{F1}=2pr/(p+r)$, and $\overline{\mathrm{IoU}}$ is averaged over all matched pairs. 
The final accuracy reward is expressed as a weighted combination of these two terms:
\begin{equation}
r_{\text{acc}} = \lambda_1\mathrm{F1} + \lambda_2\overline{\mathrm{IoU}}
\label{eq:acc_reward}
\end{equation}
This formulation provides a smooth and interpretable reward signal that balances holistic detection performance with precise spatial localization, and the use of bipartite matching ensures stable gradient feedback even under multiple-object discrepancies.

\noindent\textbf{Overall Reward and Optimization Objective.}
The overall reward $R$ integrates the structural constraint from $r_{\text{format}}$ and the perceptual feedback from $r_{\text{acc}}$ into a unified training signal. 
A small portion of the reward weight is allocated to the format term to preserve syntactic consistency while allowing the model to focus primarily on visual reasoning. 
The combined objective is formulated as:
\begin{equation}
r = (1-\alpha) r_{\text{acc}} + \alpha r_{\text{format}},
\label{eq:overall_reward}
\end{equation}
where $\alpha$ controls the relative contribution of structural validity. 
This hierarchical design ensures that the model first produces responses in a valid format before being rewarded for fine-grained detection and localization accuracy.

\paragraph{Policy Optimization.}
We optimize the policy $\pi_{\theta}$ using the Group Relative Policy Optimization (GRPO)~\cite{shao2024deepseekmath} framework, which refines multimodal generation through group-based relative rewards. 
For each input $(I_a, I_b, P)$, the model samples $G$ candidate responses $\{O^{(g)}\}_{g=1}^{G}$, each evaluated by the reward function in Eq.~\eqref{eq:overall_reward}. 
The normalized group-level advantage is defined as
\begin{equation}
A_{g} = 
\frac{r_{g} - \mathrm{mean}(\{r_{i}\}_{i=1}^{G})}
{\mathrm{std}(\{r_{i}\}_{i=1}^{G})},
\end{equation}
and the token-level advantage $A_{g,t}$ is uniformly assigned to each token within $O^{(g)}$. 
The token-level clipped surrogate objective is formulated as
\begin{equation}
\mathcal{L}^{\text{CLIP}}_{g,t}(\theta) =
\min\!\left(
r_{g,t}A_{g,t},\;
\operatorname{clip}(r_{g,t},\,1-\varepsilon,\,1+\varepsilon)\,A_{g,t}
\right),
\end{equation}
where 
$r_{g,t} =
\frac{\pi_{\theta}(o^{(g)}_{t}|o^{(g)}_{<t},I_a,I_b,P)}
{\pi_{\text{old}}(o^{(g)}_{t}|o^{(g)}_{<t},I_a,I_b,P)}$
denotes the likelihood ratio for token $t$. 
The full GRPO loss integrates the expected clipped objective with a KL-divergence regularization term:
\begin{equation}
\begin{split}
\mathcal{L}_{\text{GRPO}}(\theta)
&=  -\,\mathbb{E}_{O \sim \pi_{\theta}}
\!\left[
\frac{1}{G}\sum_{g=1}^{G}
\frac{1}{|o^{(g)}|}\sum_{t=1}^{|o^{(g)}|}
\mathcal{L}^{\text{CLIP}}_{g,t}(\theta)
\right] \\
& +\, \beta\, D_{\mathrm{KL}}\!\big(
\pi_{\theta}(\cdot\!\mid\!I_a,I_b,P)
\,\|\,
\pi_{\text{ref}}(\cdot\!\mid\!I_a,I_b,P)
\big),
\end{split}
\label{eq:grpo_loss}
\end{equation}
where $\beta$ controls the trade-off between policy updates and regularization.
The $\operatorname{clip}$ operator constrains excessive updates for stability, while the KL term penalizes deviation from the reference policy. 
Following~\cite{shao2024deepseekmath}, the KL regularization can be relaxed during early training to encourage exploration.

% To optimize the policy $\pi_{\theta}$, we adopt the GRPO~\cite{shao2024deepseekmath} framework, which refines multimodal generation through group-based relative rewards. 
% For each input $(I_a, I_b, P)$, the model samples $G$ candidate responses $\{O^{(g)}\}_{g=1}^{G}$, each evaluated by the reward function in Eq.~\eqref{eq:overall_reward}. 
% The advantage of a candidate is computed as its deviation from the group mean:
% \begin{equation}
% A(O^{(g)}) = r_{g} - \frac{1}{G}\sum_{i=1}^{G}r_{i}.
% \end{equation}
% The GRPO loss encourages higher-reward generations while maintaining stability through policy regularization:
% \begin{equation}
% \mathcal{L}_{\text{GRPO}}(\theta) = - \mathbb{E}_{\mathcal{O} \sim \pi_{\theta}} \left[ \frac{1}{G} \sum_{g=1}^{G} \frac{1}{T_g} \sum_{t=1}^{T_g} \mathcal{L}^{\text{CLIP}}_{g,t}(\theta) \right] + \beta D_{\text{KL}}(\pi_{\theta}(\cdot|I_a, I_b, P) \| \pi_{\text{ref}}(\cdot|I_a, I_b, P)) \tag{5}
% \end{equation}
% where $\pi_{\text{ref}}$ is a frozen reference policy and $\beta$ controls the KL regularization strength. 
% This objective aligns the model’s open-ended predictions with the fine-grained visual perception goals of the DiG task while ensuring stable optimization dynamics during reinforcement learning.

\subsection{Curriculum-based Difficulty Scheduling}
\label{sec:curriculum_learning}

Direct optimization of the DiG task under reinforcement learning is challenging due to the sparsity of informative rewards. 
At the early training stage, when the model is unskilled at localizing subtle visual differences, the reward signal frequently collapses to near zero, leading to unstable updates and poor convergence. 
To address this issue, we introduce a curriculum-based difficulty scheduling strategy that progressively increases task complexity, enabling the model to acquire fine-grained perceptual capabilities in a stable and scalable manner.

The curriculum is organized into multiple training phases of increasing difficulty, as illustrated in \cref{fig:method}. 
In the initial phase, the model is trained on simple instances containing a single visual difference, which provides dense and easily interpretable feedback. 
This stage encourages the policy to learn the basic skill of difference localization and to establish a reliable mapping between visual perturbations and corresponding bounding-box predictions. 
As training proceeds, the data distribution is gradually shifted toward more complex samples containing multiple differences, promoting the development of compositional reasoning and multi-object perception. 
Finally, the model is fine-tuned on a mixed-difference dataset, allowing it to generalize across diverse configurations and to handle arbitrary numbers and types of discrepancies.
During training on mixed-difference dataset, we abstain from providing the count of discrepancies for each image pair, tasking the model with inferring this information. 
This design choice elevates the task's difficulty, offering greater potential for the model's capability enhancement.

This progressive scheduling substantially stabilizes reinforcement learning by shaping the reward landscape from sparse to dense supervision, allowing the policy to acquire fine-grained perceptual skills through gradually structured visual comparisons.

\begin{table*}[t]
\centering
\setlength{\abovecaptionskip}{0.1cm}
\setlength{\belowcaptionskip}{0.1cm}
\caption{\textbf{Performance on perception benchmarks.}
Comparison across various multimodal perception benchmarks including HalBench, HRB8K, POPE, V*, VSR, CV-Bench, and MMVP. Integrating DiG consistently improves fine-grained perception and spatial reasoning for both Qwen3-VL-4B and Qwen3-VL-8B models over their respective base counterparts.}
\small
\setlength{\tabcolsep}{3mm}
\resizebox{\textwidth}{!}{
\begin{tabular}{lcccccccc}
\toprule
% \rowcolor{lightgray}
\textbf{Model} & \textbf{HalBench} & \textbf{HRB8K} & \textbf{POPE} & \textbf{V*} & \textbf{VSR} & \textbf{CV-Bench} & \textbf{MMVP} & \textbf{AVG} \\
\midrule
\textcolor{graytext}{Qwen2.5-VL-7B~\cite{bai2025qwen25vltechnicalreport}}  &  50.1 & 65.3 & 86.4 & 77.0& 77.7 & 64.9 & 54.7 &  68.0\\
\textcolor{graytext}{Qwen2.5-VL-72B~\cite{bai2025qwen25vltechnicalreport}} & 55.2& 77.1 & -- & 85.9 & -- & 83.0 & 81.7&--\\
\textcolor{graytext}{InternVL2.5-78B~\cite{chen2025expanding}} &  57.1 & 72.5 & 90.8 &  75.9 & -- & -- & 83.0 & -- \\
\midrule

% \rowcolor{yellowhighlight}
Qwen3-VL-4B-Thinking & 70.1 & 69.3 & 88.3 & 79.1 & 82.5 & 82.8 & 80.0 & 78.9\\
\rowcolor{greenhighlight}
\quad + DiG (Ours) & 73.8{\color{red}$^{\uparrow3.8}$} & 71.0{\color{red}$^{\uparrow1.7}$} & 88.5{\color{red}$^{\uparrow0.2}$} & 79.1 & 83.9{\color{red}$^{\uparrow1.4}$} & 83.8{\color{red}$^{\uparrow1.0}$} & 79.0{\color{green}$^{\downarrow1.0}$} & 79.9{\color{red}$^{\uparrow1.0}$}\\
% \rowcolor{yellowhighlight}
Qwen3-VL-8B-Thinking & 73.3 & 69.9 & 87.9 & 79.1 & 82.6 & 85.0 & 77.3 & 79.3 \\
\rowcolor{greenhighlight}
\quad + DiG (Ours) & 76.7{\color{red}$^{\uparrow3.4}$} & 70.4{\color{red}$^{\uparrow0.5}$} & 88.0{\color{red}$^{\uparrow0.1}$} & 81.2{\color{red}$^{\uparrow2.1}$} & 83.0{\color{red}$^{\uparrow0.4}$} & 85.8{\color{red}$^{\uparrow0.8}$} & 78.7{\color{red}$^{\uparrow1.4}$} & 80.5{\color{red}$^{\uparrow1.2}$} \\
\bottomrule
\end{tabular}}
\label{tab:perception}
\end{table*}

\begin{table*}[t] 
\centering 
\setlength{\abovecaptionskip}{0.1cm}
\setlength{\belowcaptionskip}{0.1cm}
\caption{\textbf{Performance on general multimodal benchmarks.}
Evaluation on standard multimodal benchmarks including MMBench, MMVet, MMStar, SQA$^I$, VQA$^T$, MME, and AI2D. Incorporating DiG into Qwen3-VL models consistently enhances overall reasoning and perception capabilities, demonstrating strong transferability beyond fine-grained tasks.} 
\setlength{\tabcolsep}{3mm}
\resizebox{\textwidth}{!}{
\begin{tabular}{lccccccccc} 
\toprule

\textbf{Model} & \textbf{LLM} & \textbf{MMBench} & \textbf{MMVet} & \textbf{MMStar} & \textbf{SQA$^I$} & \textbf{VQA$^T$}  & \textbf{MME} & \textbf{AI2D} \\
\midrule
\textcolor{graytext}{LLaVA-NeXT~\cite{liu2024llavanext}} & Vicuna1.5-7B & 67.4 & 43.9 & 37.7 & 70.1 & 64.9 & 1519& 66.6\\
\textcolor{graytext}{InternVL3.5-38B~\cite{wang2025internvl3}} & Qwen3-32B & 87.3 & 82.2 & 75.3 & -- & 82.7 & --& 87.8\\
\textcolor{graytext}{Qwen2.5-VL 72B~\cite{bai2025qwen25vltechnicalreport}} & Qwen2.5-72B & 88.6 & 76.2  & 70.8 &-- & 84.9 & --& 83.9\\
\midrule
Qwen3-VL & Qwen3-4B & 83.7 & 76.0 & 66.7 & 86.9 & 76.2 & 1592.4 & 81.0 \\
\rowcolor{greenhighlight}
\quad + DiG (Ours) & Qwen3-4B & 84.5{\color{red}$^{\uparrow0.8}$} & 77.3{\color{red}$^{\uparrow1.3}$} & 70.2{\color{red}$^{\uparrow3.5}$} & 89.2{\color{red}$^{\uparrow2.3}$} & 76.5{\color{red}$^{\uparrow0.3}$} & 1643.4{\color{red}$^{\uparrow51.0}$} &  82.2{\color{red}$^{\uparrow1.2}$}\\
Qwen3-VL & Qwen3-8B & 85.5 & 76.4 & 71.7 & 90.3 & 75.5 & 1648.4 &  82.5 \\
\rowcolor{greenhighlight}
\quad + DiG (Ours) & Qwen3-8B & 87.2{\color{red}$^{\uparrow2.2}$} & 76.8{\color{red}$^{\uparrow0.4}$} & 72.7{\color{red}$^{\uparrow1.0}$} & 90.5{\color{red}$^{\uparrow0.2}$} & 77.5{\color{red}$^{\uparrow2.0}$} & 1665.9{\color{red}$^{\uparrow17.5}$} &  84.1{\color{red}$^{\uparrow1.6}$}\\
\bottomrule 
\end{tabular}
}
\label{tab:general}
\end{table*}

\section{Experiments}
\subsection{Implementation Details}

\paragraph{Training Settings.}
All experiments are conducted on the proposed Differential Grounding (DiG) dataset, generated using the 3D rendering pipeline described in \cref{sec:data_construction}. 
The training corpus comprises approximately 4.8K image pairs, split evenly across three subsets: single-difference, double-difference, and mixed-difference (about 1.6K each). Mixed-difference scenes contain up to four visual discrepancies.
We adopt Qwen3-VL-8B-Thinking and Qwen3-VL-4B-Thinking~\cite{bai2025qwen3} as multimodal backbones, and perform reinforcement post-training using the EasyR1~\cite{zheng2025easyr1,sheng2025hybridflow} framework with the GRPO objective and KL-regularized policy optimization. 
Training follows the curriculum strategy detailed in \cref{sec:curriculum_learning}, gradually increasing task complexity from simple to mixed-difference instances. 
All implementation and hyperparameter details are provided in the Appendix.

\paragraph{Evaluation Setup.}
We evaluate the proposed approach on a comprehensive suite of multimodal benchmarks covering visual perception, grounding, and general understanding. 
The perception evaluation includes HalBench~\cite{guan2024hallusionbench}, HRBench-8K~\cite{wang2025divide}, POPE~\cite{li2023evaluating}, V*~\cite{wu2023vguidedvisualsearch}, VSR~\cite{liu2023visual}, CV-Bench~\cite{tong2024cambrian}, and MMVP~\cite{tong2024eyes}. 
Grounding performance is assessed on RefCOCO, RefCOCO+~\cite{yu2016modeling,kazemzadeh-etal-2014-referitgame}, and RefCOCOg~\cite{mao2016generation}, while broader multimodal reasoning and comprehension are examined on MMBench~\cite{liu2024mmbench}, MM-Vet~\cite{yu2023mm}, MMStar~\cite{chen2024we}, ScienceQA~\cite{lu2022learn}, TextVQA~\cite{singh2019towards}, MME~\cite{fu2023mme}, and AI2D~\cite{kembhavi2016diagram}. 
All evaluations follow the standard protocols of each benchmark to ensure fairness and reproducibility, with detailed settings provided in the Appendix.

\subsection{Main Results}

\paragraph{Performance on Perception Benchmarks.}

As shown in \cref{tab:perception}, DiG consistently improves both the 4B and 8B variants of Qwen3-VL-Thinking across a broad range of perception-oriented benchmarks. 
Notably, the 8B model achieves substantial gains of +3.4 on HalBench and +2.1 on V*, accompanied by steady improvements on VSR and CV-Bench. 
The 4B variant follows a similar trend, yielding up to +3.8 improvement on HalBench and noticeable gains on HRB8K and VSR, confirming that the proposed differential grounding generalizes effectively across different model scales. 
These consistent enhancements indicate that DiG strengthens the model’s sensitivity to fine-grained visual cues and its ability to maintain spatial consistency under complex visual perturbations, leading to reduced hallucination and enhanced perceptual fidelity. 
Taken together, these results highlight that differential grounding provides a scalable and principled way to reinforce fine-grained perceptual understanding and bridge structured perception with higher-level multimodal reasoning.

\paragraph{Performance on General Multimodal Benchmarks.}
We further evaluate DiG on a suite of comprehensive multimodal benchmarks that jointly assess holistic perception and reasoning ability, as summarized in \cref{tab:general}. 
Across all evaluated tasks, DiG delivers consistent performance gains for both the 4B and 8B variants of Qwen3-VL, demonstrating the generality and robustness of the proposed reinforcement formulation. 
The 4B model achieves notable improvements of +3.5 on MMStar and +2.3 on ScienceQA, while the 8B model exhibits steady gains on MMBench (+2.2) and MME (+17.5), indicating that differential grounding effectively strengthens multimodal reasoning and structured visual-text alignment. 
Such enhancements are particularly remarkable given that DiG-equipped models, despite their relatively smaller scale, achieve performance comparable to or exceeding that of several larger proprietary systems reported in prior literature. 
This suggests that the improvements brought by DiG stem from more efficient representation learning and better reward-driven perception refinement rather than model size alone. 
Overall, these results confirm that DiG fosters more structured multimodal understanding, enabling models to better integrate visual perception with abstract reasoning under a unified reinforcement framework. 
\looseness=-1

\begin{table*}[t]
\centering
\setlength{\abovecaptionskip}{0.05cm}
\setlength{\belowcaptionskip}{0.05cm}
\small
\caption{Evaluation on visual grounding benchmarks.
Results on referring expression comprehension datasets RefCOCO, RefCOCO+, and RefCOCOg. Incorporating DiG into Qwen3-VL models consistently improves localization accuracy across different IoU thresholds, demonstrating stronger fine-grained spatial grounding and generalization ability. }
\setlength{\tabcolsep}{1.5mm}
\resizebox{\textwidth}{!}{
\begin{tabular}{lccccccccccccc}
\toprule
% \rowcolor{gray!20}
\multicolumn{14}{c}{\textbf{RefCOCO}}\\
\midrule
\textbf{method} & \textbf{size} & val\textsubscript{@50} & testA\textsubscript{@50} & testB\textsubscript{@50} & val\textsubscript{@75} & testA\textsubscript{@75} & testB\textsubscript{@75} & val\textsubscript{@95} & testA\textsubscript{@95} & testB\textsubscript{@95} & val\textsubscript{avg} & testA\textsubscript{avg} & testB\textsubscript{avg} \\
\cmidrule(lr){1-2} \cmidrule(lr){3-5} \cmidrule(lr){6-8} \cmidrule(lr){9-11} \cmidrule(lr){12-14}
% \textcolor{graytext}{MDETR [25]} & - & 87.5 & 90.4 & 82.6 & - & - & - & - & - & - & - & - & - \\
% \textcolor{graytext}{OFA [62]} & - & 88.4 & 90.4 & 83.3 & - & - & - & - & - & - & - & - & - \\
% \midrule
Qwen3-VL  & 4B & 85.7 & 90.1 & 79.1 & 77.6 & 81.9 & 69.0 & 28.7 & 30.8 & 23.1 & 64.0 & 67.6 & 57.1 \\
\rowcolor{greenhighlight}
\quad + DiG (Ours) & 4B & 88.6 & 91.5 & 83.5 & 80.3 & 83.5 & 72.3 & 31.6 & 35.2 & 26.6 & 66.8 & 70.1 & 60.8 \\
\midrule
Qwen3-VL  & 8B & 86.9 & 90.6 & 81.7 & 78.9 & 82.5 & 70.9 & 29.1 & 31.3 & 23.8 & 65.0 & 68.1 & 58.8 \\
\rowcolor{greenhighlight}
\quad + DiG (Ours) & 8B& 90.0 & 91.9 & 86.4 & 81.0 & 84.0 & 74.9 & 29.9 & 33.9 & 25.3 & 67.0 & 69.9 & 62.2 \\
\midrule

% \rowcolor{gray!20}
\multicolumn{14}{c}{\textbf{RefCOCO+}}\\
\midrule
\textbf{method} & \textbf{size} & val\textsubscript{@50} & testA\textsubscript{@50} & testB\textsubscript{@50} & val\textsubscript{@75} & testA\textsubscript{@75} & testB\textsubscript{@75} & val\textsubscript{@95} & testA\textsubscript{@95} & testB\textsubscript{@95} & val\textsubscript{avg} & testA\textsubscript{avg} & testB\textsubscript{avg} \\
\cmidrule(lr){1-2} \cmidrule(lr){3-5} \cmidrule(lr){6-8} \cmidrule(lr){9-11} \cmidrule(lr){12-14}
% \textcolor{graytext}{MDETR [25]} & - & 81.1 & 85.5 & 72.9 & - & - & - & - & - & - & - & - & - \\
% \textcolor{graytext}{OFA [62]} & - & 81.3 & 87.1 & 74.2 & - & - & - & - & - & - & - & - & - \\
% \midrule
Qwen3-VL  & 4B & 80.3 & 87.0 & 72.9 & 72.8 & 78.8 & 63.6 & 27.0 & 29.6 & 21.7 & 60.0 & 65.1 & 52.7 \\
\rowcolor{greenhighlight}
\quad + DiG (Ours) & 4B & 82.9 & 87.4 & 76.4 & 75.2 & 79.3 & 66.3 & 29.9 & 33.8 & 24.3 & 62.6 & 66.8 & 55.7 \\
\midrule
Qwen3-VL  & 8B & 81.8 & 86.6 & 73.9 & 74.0 & 78.5 & 64.6 & 27.9 & 30.9 & 21.7 & 61.2 & 65.3 & 53.4 \\
\rowcolor{greenhighlight}
\quad + DiG (Ours) & 8B & 84.2 & 87.7 & 77.6 & 76.1 & 80.1 & 68.1 & 28.6 & 32.2 & 23.5 & 63.0 & 66.7 & 56.4\\
\midrule

% \rowcolor{gray!20}
\multicolumn{14}{c}{\textbf{RefCOCOg}}\\
\midrule
\textbf{method} & \textbf{size} & val\textsubscript{@50} & test\textsubscript{@50} & & val\textsubscript{@75} & test\textsubscript{@75} &  & val\textsubscript{@95} & test\textsubscript{@95} &  & val\textsubscript{avg} & test\textsubscript{avg} & \\
\cmidrule(lr){1-2} \cmidrule(lr){3-5} \cmidrule(lr){6-8} \cmidrule(lr){9-11} \cmidrule(lr){12-14}
% \textcolor{graytext}{MDETR [25]} & - & 83.3 & 83.3 &  & - & - &  & - & - &  & - & - &  \\
% \textcolor{graytext}{OFA [62]} & - & 82.2 & 82.3 &  & - & - &  & - & - &  & - & - & \\
% \midrule
Qwen3-VL  & 4B & 85.0 & 85.0 &  & 75.2 & 76.1 & & 27.1 & 27.6 &  & 62.4 & 62.9 &  \\
\rowcolor{greenhighlight}
\quad + DiG (Ours) & 4B & 87.2 & 87.2 &  & 77.8 & 78.2 &  & 30.7 & 31.2 &  & 65.3 & 65.6 & \\
\midrule
Qwen3-VL  & 8B & 86.3 & 86.5 &  & 76.1 & 77.0 &  & 26.3 & 27.4 &  & 62.9 & 63.6 & \\
\rowcolor{greenhighlight}
\quad + DiG (Ours) & 8B & 87.7 & 88.0 &  & 77.7 & 78.5 &  & 28.8 & 29.8 &  & 64.7 & 65.4 & \\
\bottomrule
\end{tabular}
}
\label{tab:grounding}
\vspace{-0.2cm}
\end{table*}

\paragraph{Performance on Grounding Benchmarks.}

% \cref{tab:grounding} summarizes the results on the RefCOCO, RefCOCO+, and RefCOCOg benchmarks, which evaluate fine-grained visual grounding under varying linguistic and spatial conditions. 
% Models trained with DiG consistently outperform their baselines across all settings. 
% Both the 4B and 8B variants exhibit an average improvement of approximately 2--3 points across all datasets, confirming that the proposed differential grounding task substantially enhances fine-grained perceptual alignment and spatial reasoning ability. 
% Such improvement may also partly stem from employing a grounding-oriented output format during reinforcement optimization, where format-consistent supervision encourages more stable visual-text alignment and further reinforces domain-specific multimodal capability.

\cref{tab:grounding} summarizes the results on the RefCOCO, RefCOCO+, and RefCOCOg benchmarks, which evaluate fine-grained visual grounding under varying linguistic and spatial conditions. 
Across all datasets and evaluation splits, models trained with DiG consistently outperform their base counterparts, demonstrating the effectiveness of differential grounding in improving region-level perception. 
Specifically, the 8B variant achieves average improvements of approximately 2--3 points across all benchmarks, with particularly strong gains on RefCOCO (+2.0--+3.4) and RefCOCO+ (+1.4--+3.0), while the 4B model exhibits comparable trends. 
These results indicate that DiG substantially enhances spatial reasoning and fine-grained localization, enabling more accurate alignment between referring expressions and visual regions. 
We attribute this improvement not only to the reinforcement of object-level perceptual precision, but also to the grounding-oriented output format adopted during optimization, which provides consistent supervision and stabilizes multimodal alignment during policy refinement.

\subsection{Ablation Study and Analysis}
\begin{figure}[t]
  \centering
   \setlength{\abovecaptionskip}{0.cm}
   \setlength{\belowcaptionskip}{0.cm}
   \includegraphics[width=\linewidth]{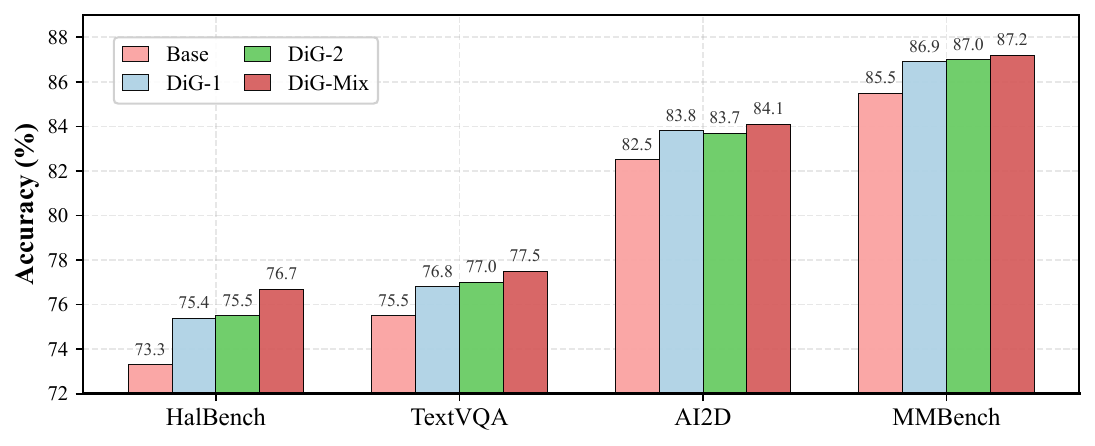}
   \caption{Ablation on curriculum scheduling in DiG training.
Progressively increasing task complexity from single-difference (DiG-1) to double-difference (DiG-2) and mixed-difference (DiG-Mix) leads to consistent performance gains across HalBench, TextVQA, AI2D, and MMBench.}
   \label{fig:ablation}
   \vspace{-0.2cm}
\end{figure}

\paragraph{Ablation on Curriculum Scheduling.}
To evaluate the effectiveness of the proposed curriculum-based optimization, we conduct an ablation study across different stages of the DiG training pipeline.
As shown in \cref{fig:ablation}, the model trained on the single-difference stage (DiG-1) already achieves clear gains over the base model, indicating that fine-grained discrepancy supervision effectively enhances perceptual alignment.
Building upon DiG-1, further training on the double-difference stage (DiG-2) yields additional gains on most benchmarks, indicating the benefit of progressively increasing perceptual complexity.
Finally, fine-tuning on the mixed-difficulty data (DiG-Mix) leads to the best overall performance, confirming that progressive exposure to increasingly complex visual discrepancies yields more robust and generalizable visual understanding.

\paragraph{Ablation on Reward.}
\begin{table}[t]
\centering
\setlength{\abovecaptionskip}{0.cm}
\setlength{\belowcaptionskip}{0.cm}
\caption{Ablation on reward components.
Evaluating different reward terms in DiG shows that combining spatial (IoU) and detection (F1) signals yields the best overall performance across RefCOCO, HalBench, MMBench, and AI2D.}
\resizebox{\linewidth}{!}{
\begin{tabular}{cccccccccc}
\toprule
\multirow{2}{*}{\textbf{Format}} & \multirow{2}{*}{\textbf{IoU}} & \multirow{2}{*}{\textbf{F1}} & \multicolumn{3}{c}{\textbf{RefCOCO}} & \multirow{2}{*}{\textbf{HalBench}} & \multirow{2}{*}{\textbf{MMB}} & \multirow{2}{*}{\textbf{AI2D}}\\
\cmidrule(lr){4-6}
 & & & val\textsubscript{@50} & testA\textsubscript{@50} & testB\textsubscript{@50} & & \\
\midrule
\rowcolor{gray!20}
\cmark & \xmark & \xmark & -- & -- & -- & -- & -- & -- \\
\cmark & \cmark & \xmark & 84.2 & 89.3 & 76.7 & 70.8 & 83.2 & 81.7\\
\cmark & \xmark & \cmark & 87.8 & 91.1 & 82.4 & 72.8 & 84.4 & 82.1\\
\cmark & \cmark & \cmark & 88.6 & 91.5 & 83.5 & 73.8 & 84.5 & 82.2\\
\bottomrule
\end{tabular}
}
\label{tab:reward_ablation}
\vspace{-0.5cm}
\end{table}

We further analyze the effect of different reward components by selectively activating the IoU and F1 terms while keeping the format reward enabled (\cref{tab:reward_ablation}). 
Relying solely on IoU leads to a noticeable performance drop, suggesting that spatial overlap alone does not sufficiently capture task success, as the model can obtain relatively high IoU scores without correctly identifying all target regions. 
In contrast, using only the F1-based reward improves precision by emphasizing detection completeness but still lacks stable spatial grounding. 
Integrating both IoU and F1 yields the best overall performance, achieving 88.6/91.5/83.5 on RefCOCO and consistent gains across HalBench, MMBench, and AI2D. 
These findings demonstrate that a balanced combination of spatial and categorical signals is essential for shaping an informative reward landscape and guiding the model toward holistic grounding behavior.

% \subsection{Analysis}

\begin{figure}[h]
  \centering
   \setlength{\abovecaptionskip}{0.cm}
   \setlength{\belowcaptionskip}{0.cm}
   \includegraphics[width=\linewidth]{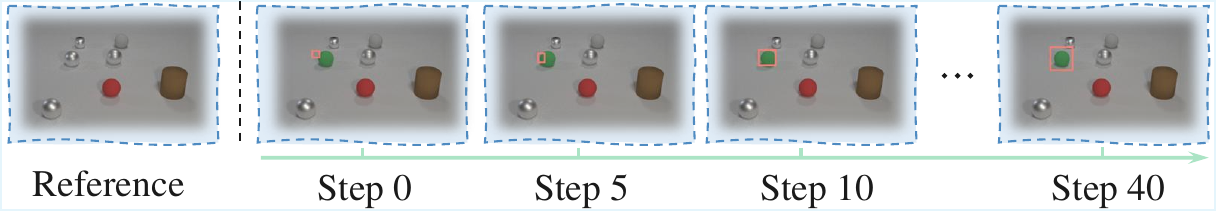}
   \caption{Visualization of differential grounding dynamics.
The model progressively refines discrepancy localization from coarse initial predictions to precise detections, revealing the emergence of fine-grained perceptual alignment during reinforcement learning.}
   \label{fig:case}
   \vspace{-0.3cm}
\end{figure}

\paragraph{Case Study on Differential Grounding Dynamics.}
To better understand the learning behavior of our model, we visualize its intermediate predictions during reinforcement optimization on a validation sample, as shown in \cref{fig:case}. 
The model progressively refines its localization of visual discrepancies, evolving from coarse and partially incorrect bounding boxes at early steps to precise and consistent detections after convergence. 
This gradual improvement reflects the emergence of fine-grained perceptual alignment between the reference and modified images, indicating that the model gradually internalizes the notion of differential grounding through reinforcement feedback.

\begin{figure}[h]
  \centering
   \setlength{\abovecaptionskip}{0.cm}
   \setlength{\belowcaptionskip}{0.cm}
  \begin{subfigure}{0.32\linewidth}
    \includegraphics[width=\linewidth]{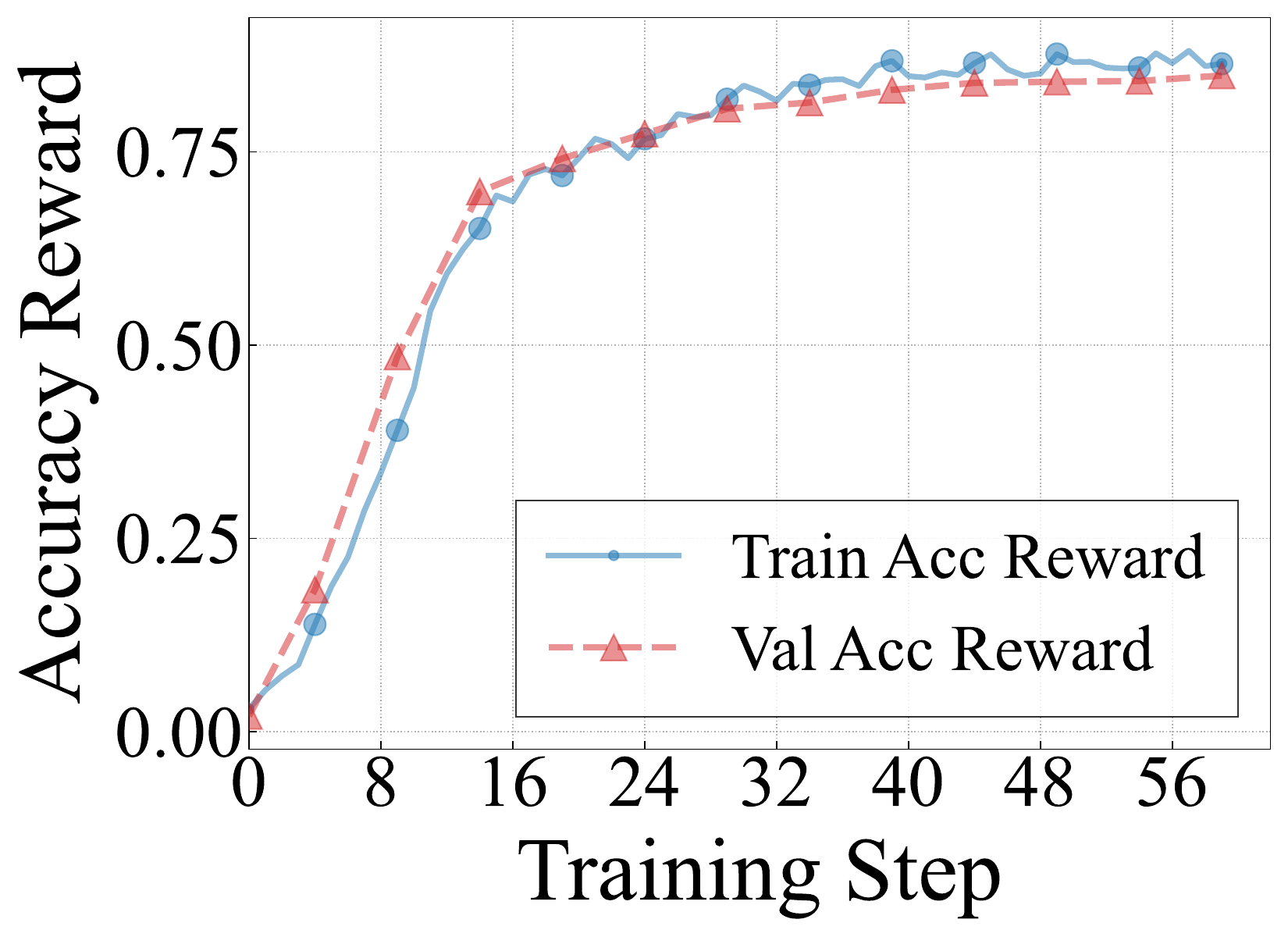}
    \caption{DiG-1}
    \label{fig:reward-fd1}
  \end{subfigure}
  \hfill
  \begin{subfigure}{0.32\linewidth}
    \includegraphics[width=\linewidth]{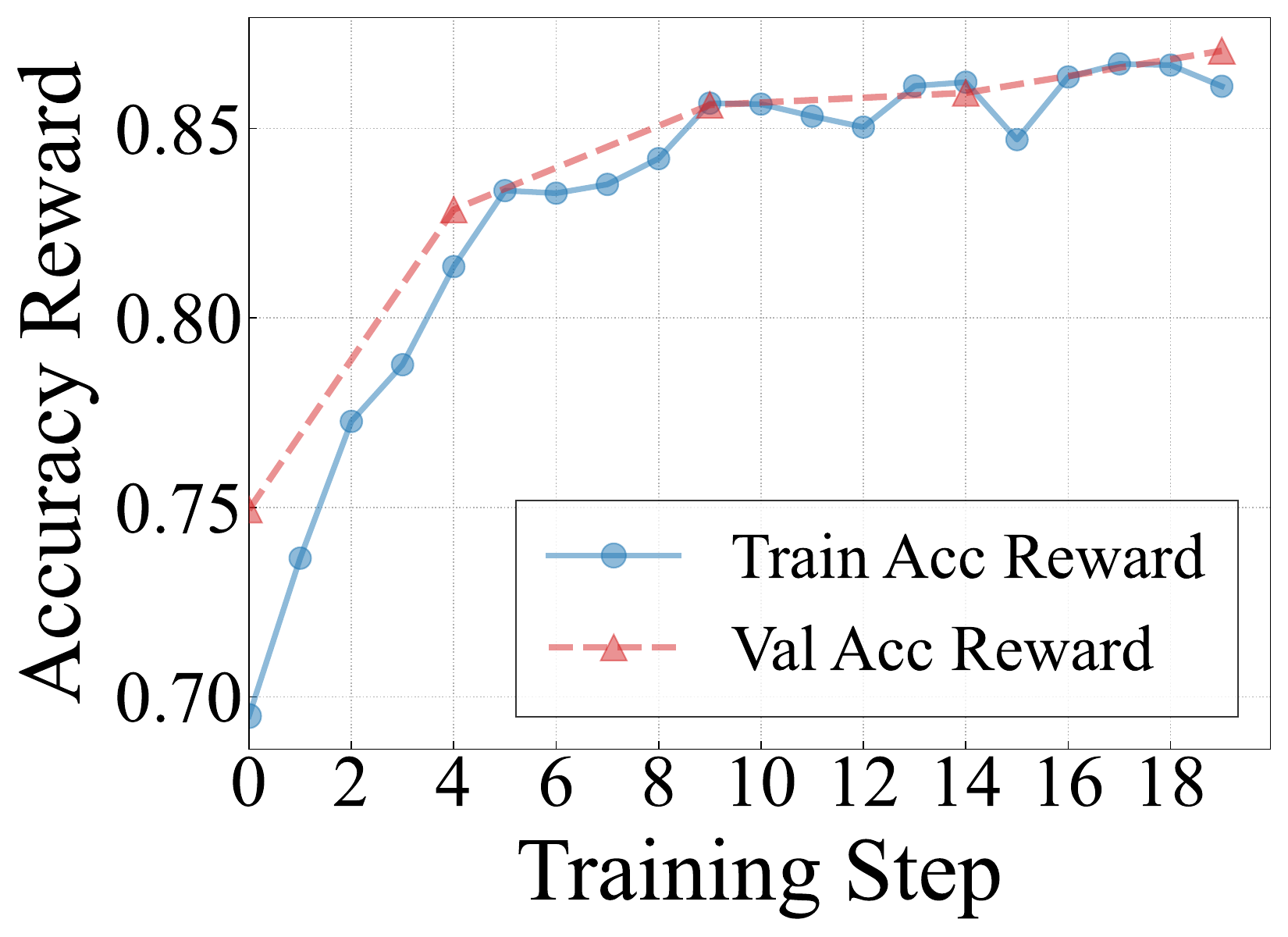}
    \caption{DiG-2}
    \label{fig:reward-fd2}
  \end{subfigure}
  \hfill
  \begin{subfigure}{0.32\linewidth}
    \includegraphics[width=\linewidth]{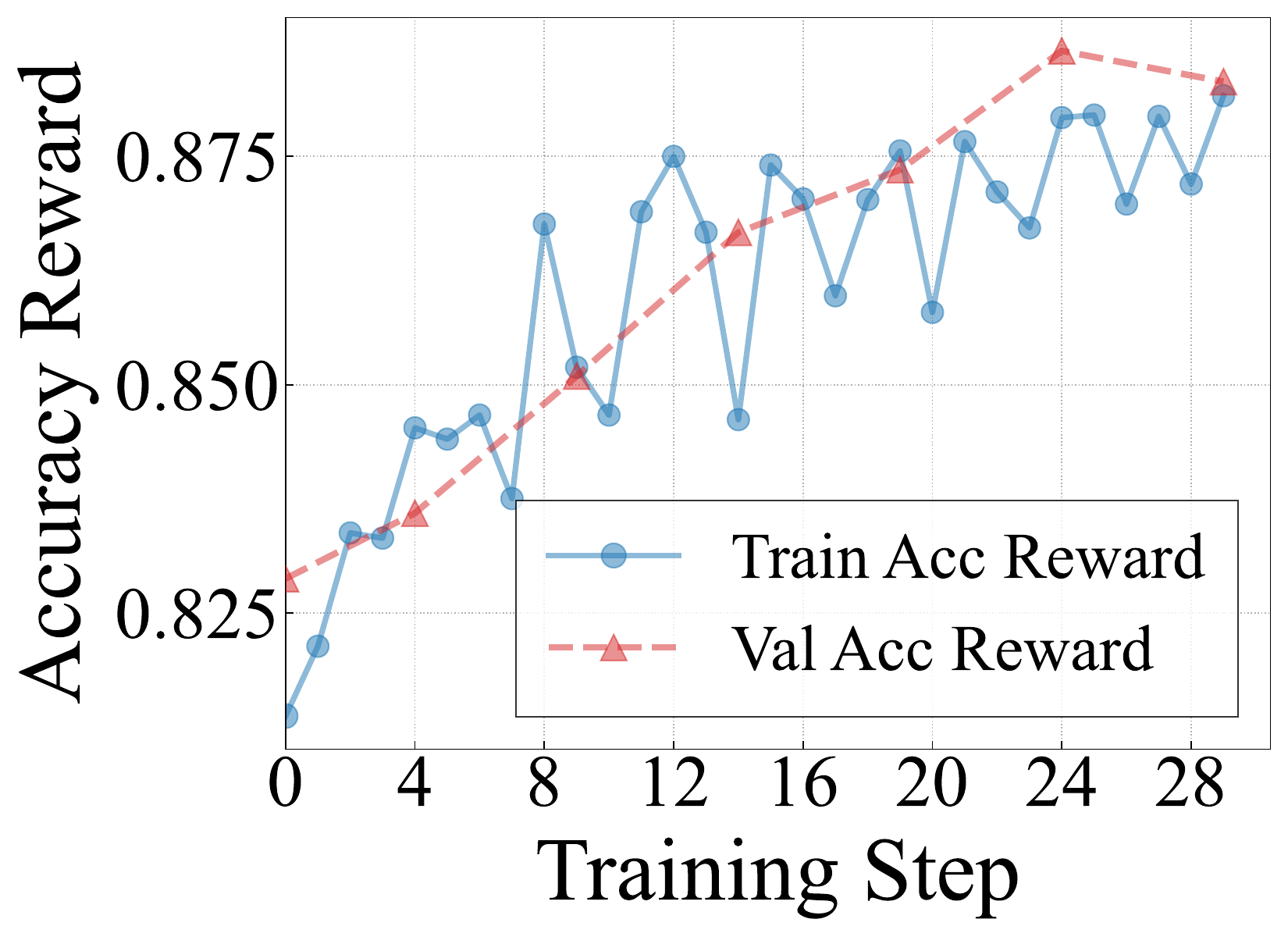}
    \caption{DiG-mix}
    \label{fig:reward-mix}
  \end{subfigure}
  \caption{Accuracy reward across curriculum stages of DiG.}
  \label{fig:reward_curve}
  \vspace{-0.4cm}
\end{figure}

\paragraph{Curriculum Optimization Behavior.}
We visualize the accuracy reward during reinforcement learning across the three curriculum stages in \cref{fig:reward_curve}. 
As shown in the figure, the model initially struggles to obtain meaningful rewards, reflecting the difficulty of optimizing fine-grained perception from sparse feedback. 
In the single-difference stage (DiG-1), the reward rises rapidly as the model learns to detect and localize basic visual discrepancies. 
Subsequent training on the double-difference stage (DiG-2) further refines spatial reasoning and stabilizes convergence. 
Finally, the mixed-difficulty stage (DiG-Mix) maintains steady improvements under increased task complexity, indicating that curriculum scheduling effectively mitigates reward sparsity and facilitates robust policy optimization.

\begin{figure}[h]
  \centering
   \setlength{\abovecaptionskip}{0.cm}
   \setlength{\belowcaptionskip}{0.cm}
   \includegraphics[width=\linewidth]{figures/vis_case.pdf}
   \caption{Case study on fine-grained visual reasoning.
Our model yields more accurate responses than Qwen3-VL, particularly for subtle spatial and attribute-based questions.}
   \label{fig:case_vis}
   \vspace{-0.5cm}
\end{figure}

\paragraph{Qualitative Analysis.}
We further present qualitative examples in \cref{fig:case_vis} to illustrate the effect of DiG on fine-grained visual reasoning. 
Compared with Qwen3-VL, our model produces more accurate answers for queries involving subtle spatial relations and object-level attributes. 
In the first two cases, the baseline fails to correctly judge relative size and depth ordering, while our model provides consistent and spatially coherent predictions. 
Similarly, in the lower examples, our model accurately recognizes fine attribute details such as color and object category that are easily overlooked by the baseline. 
These results highlight that DiG enhances the model’s sensitivity to local visual discrepancies and strengthens its capacity for precise perceptual reasoning beyond coarse semantic understanding.
\section{Conclusion}

In this work, we introduced DiG (Differential Grounding), a novel proxy framework designed to enhance fine-grained visual perception and spatial reasoning in multimodal large language models. 
By framing differential comparison between paired images as a grounding task, DiG encourages models to develop precise visual discrimination and object-level awareness beyond conventional semantic alignment. 
Our automated 3D rendering pipeline enables scalable data synthesis with controllable discrepancies, while the curriculum-based reinforcement learning strategy effectively mitigates reward sparsity and stabilizes optimization. 
Extensive experiments across perception, grounding, and general benchmarks demonstrate that DiG consistently improves fine-grained perceptual capability and generalization across scales. 
We believe that differential grounding offers a scalable and principled paradigm for bridging high-level reasoning and low-level perception in multimodal large language models, and provides a promising direction toward more perceptually aligned multimodal intelligence.

\section*{Acknowledgements}

This research was supported by the Strategic Priority Research Program of Chinese Academy of Sciences (XDA0490000) and the National Natural Science Foundation of China (62276245).
{
    \small
    \bibliographystyle{ieeenat_fullname}
    \bibliography{main}
}

\clearpage
\setcounter{page}{1}
\maketitlesupplementary

\section{Implementation Details}

\paragraph{Training Configuration}
All experiments are conducted on the proposed Differential Grounding (DiG) dataset, which is procedurally generated through the 3D rendering pipeline introduced in Section~\ref{sec:data_construction}.
The corpus contains approximately 4.8K image pairs, evenly distributed across three subsets: single-difference, double-difference, and mixed-difference scenes (around 1.6K per subset).
Each mixed-difference scene includes up to four visual discrepancies, while every image comprises up to ten distinct objects, ensuring a diverse range of visual layouts and attribute variations.

We employ Qwen3-VL-8B-Thinking and Qwen3-VL-4B-Thinking~\cite{bai2025qwen3} as multimodal backbones.
Reinforcement post-training is carried out using the EasyR1 framework~\cite{zheng2025easyr1,sheng2025hybridflow}, optimized with the GRPO objective and KL-regularized policy optimization.
Training follows the curriculum strategy outlined in Section~\ref{sec:curriculum_learning}, progressively increasing task complexity from single- to mixed-difference instances.

All hyperparameters, training schedules, and optimization settings are listed in Table~\ref{tab:rl_config}.

\begin{table}[htbp]
\centering
\small
\setlength{\tabcolsep}{6pt}
\renewcommand{\arraystretch}{1.05}
\resizebox{\linewidth}{!}{
\begin{tabular}{lll}
\toprule
\textbf{Component} & \textbf{Parameter} & \textbf{Value} \\
\midrule
\multirow{3}{*}{\textbf{Algorithm}} 
& KL coefficient & $1.0 \times 10^{-2}$ \\
& Filter range & [0.01, 0.99] \\
& Reward weight $\alpha$ & 0.1 \\
\midrule
\multirow{5}{*}{\textbf{Optimization}} 
& Learning rate & $1.0 \times 10^{-6}$ \\
& Weight decay & $1.0 \times 10^{-2}$ \\
& Optimizer & AdamW \\
& Gradient clipping & 1.0 \\
& Warmup ratio & 0.0 \\
\midrule
\multirow{3}{*}{\textbf{Model}} 
& Gradient checkpointing & True \\
& Vision tower & Trainable \\
& FSDP sharding & Full \\
\midrule
\multirow{4}{*}{\textbf{Rollout}} 
& $n$ & 5 \\
& Temperature & 1.0 \\
& Top-p & 1.0 \\
& Validation override & $T{=}0.6$, $p{=}0.95$, $n{=}1$ \\
\midrule
\multirow{3}{*}{\textbf{Training Schedule}} 
& Stage 1 steps & 60 \\
& Stage 2 steps & 20 \\
& Stage 3 steps & 30 \\
\bottomrule
\end{tabular}}
\caption{\textbf{Reinforcement Post-Training Configuration.} 
Key hyperparameters and rollout settings across the three curriculum stages.}
\label{tab:rl_config}
\end{table}

To enable consistent instruction-following behavior during multimodal reasoning and grounding, we adopt a structured prompt format for the Differential Grounding (DiG) task. 
The prompt template we use is shown below.

\begin{tcolorbox}[
    colback=white,           % 内容背景白
    colframe=black!80,       % 边框浅灰
    colbacktitle=gray!45,    % 标题栏浅灰
    coltitle=black,          % 标题文字黑色
    fonttitle=\bfseries,     % 标题加粗
    title=Prompt for DiG Task,
    boxrule=0.7pt,           % 边框线稍细
    arc=2pt,                 % 轻微圆角
    left=5pt, right=5pt, top=4pt, bottom=4pt,
]
\small
\texttt{<image>}\texttt{<image>}You will be given two separate images.
- The first image is the 'before' version.

- The second image is the 'after' version.

Your task is to compare them and identify all 'changed regions'. Provide your answer as a JSON list of bounding boxes.
All bounding boxes you provide must be on the **'before' (first) image**.

Follow these rules:
1. **Item is Missing or Changed:** If an item from the 'before' image is **missing** or **looks different** (e.g., color, shape, state) in the 'after' image, box the **original item on the 'before' image**.
2. **Item is New:** If a **new item appears** in the 'after' image (in a previously empty spot), box the **corresponding empty area on the 'before' image**.

**Important:** All coordinates must be relative to the **first ('before') image**. Use the format [x\_1, y\_1, x\_2, y\_2] with absolute pixel coordinates.
\end{tcolorbox}

For Stage 1 and Stage 2, the model is additionally informed of the number of differences to facilitate curriculum-based reasoning.

\paragraph{Evaluation Configuration.}
Evaluation is conducted across a comprehensive suite of multimodal benchmarks, covering perception, grounding, and reasoning capabilities.
The visual perception evaluation includes HalBench~\cite{guan2024hallusionbench}, HRBench-8K~\cite{wang2025divide}, POPE~\cite{li2023evaluating}, V*~\cite{wu2023vguidedvisualsearch}, VSR~\cite{liu2023visual}, CV-Bench~\cite{tong2024cambrian}, and MMVP~\cite{tong2024eyes}.
Grounding performance is assessed on RefCOCO, RefCOCO+~\cite{yu2016modeling,kazemzadeh-etal-2014-referitgame}, and RefCOCOg~\cite{mao2016generation}.
For multimodal reasoning and comprehension, we evaluate on MMBench~\cite{liu2024mmbench}, MM-Vet~\cite{yu2023mm}, MMStar~\cite{chen2024we}, ScienceQA~\cite{lu2022learn}, TextVQA~\cite{singh2019towards}, MME~\cite{fu2023mme}, and AI2D~\cite{kembhavi2016diagram}.

Inference is conducted using the vLLM framework for efficient batched decoding across all benchmarks.
Each evaluation employs batched inference with dynamic padding and a maximum generation length of 4096 tokens.
The sampling temperature is fixed at 0, ensuring deterministic outputs across runs.
During evaluation, the model generates both intermediate reasoning traces and final answers.
For perception and general QA benchmarks, the model generates intermediate reasoning traces enclosed by the thinking tag.
Following the “Think–Then–Answer” convention, only the text after \texttt{</think>} is extracted and used as the final answer.
In contrast, for grounding tasks, predictions are taken directly from the model’s generated bounding boxes without post-hoc extraction.
This extraction rule is consistently applied across all benchmarks to ensure comparable evaluation.
All results reported in the main paper are obtained under these unified inference settings.

\section{DiG Data Construction}

To construct the Differential Grounding (DiG) dataset, we employ a controllable 3D scene generation pipeline that procedurally renders paired images under well-defined attribute configurations.  
Each scene consists of multiple objects parameterized by a set of compositional attributes, allowing precise manipulation of object-level differences across image pairs.

\paragraph{Attribute Definition.}
Objects are instantiated from four primary attribute groups: \textbf{shape}, \textbf{color}, \textbf{size}, and \textbf{material}.  
The \emph{shape} attribute includes three geometric primitives—\texttt{Cube}, \texttt{Sphere}, and \texttt{Cylinder}—which form the core set of object templates.  
The \emph{color} palette covers eight distinct hues (\texttt{blue}, \texttt{brown}, \texttt{cyan}, \texttt{gray}, \texttt{green}, \texttt{purple}, \texttt{red}, \texttt{yellow}) with fixed RGB coefficients to ensure visual consistency.  
Each object further takes one of three \emph{size} levels (\texttt{small}, \texttt{medium}, \texttt{large}) corresponding to normalized scale factors of 0.4, 0.6, and 0.8, respectively.  
Finally, the \emph{material} attribute controls surface reflectance through physically-based rendering parameters, where \texttt{metal} corresponds to $(\textit{metallic}=1.0,\, \textit{roughness}=0.2)$ and \texttt{matte} to $(\textit{metallic}=0.0,\, \textit{roughness}=0.8)$.

\begin{figure}[htbp]
  \centering
   \setlength{\abovecaptionskip}{0.cm}
   \setlength{\belowcaptionskip}{0.cm}
   \includegraphics[width=\linewidth]{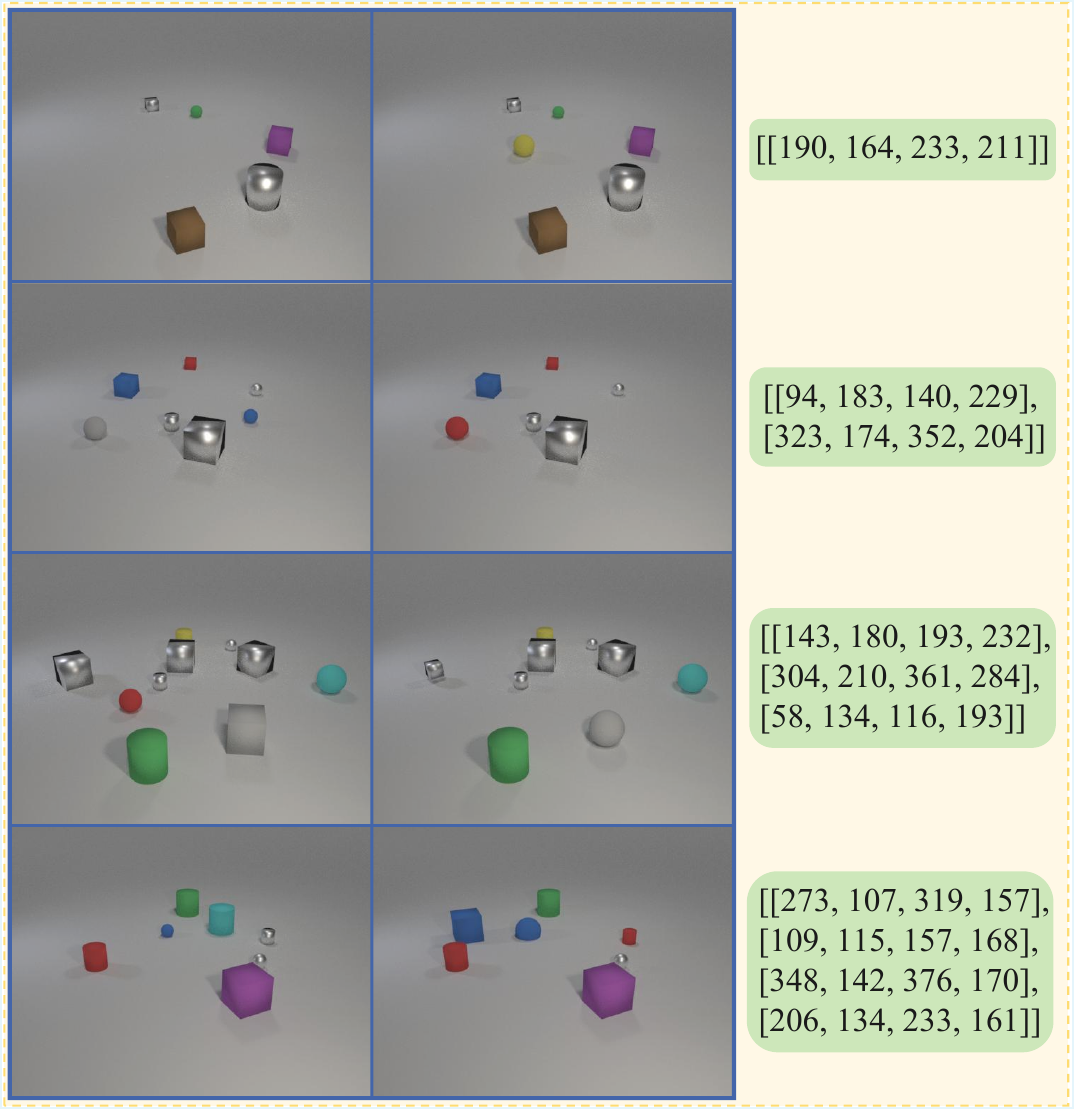}
   \caption{
\textbf{Examples of synthesized Differential Grounding (DiG) scenes.} 
Each row shows an image pair with controlled visual differences in shape, color, size, or material, and the corresponding bounding boxes on the \emph{“before”} image. }
   \label{fig:dig_examples}
\end{figure}

\paragraph{Scene Composition.}
For each scene, we randomly sample a subset of objects from the attribute space to populate a 3D layout.  
Controlled perturbations are then applied to introduce visual differences between paired images, such as changes in object shape, color, size, material, or count.  
This design enables a rich combination space while maintaining interpretable and disentangled factors of variation.  
The entire generation process is implemented directly in Python using Blender’s rendering API, allowing precise reproducibility and randomized yet structured diversity across instances.
Representative examples of the synthesized image pairs are illustrated in Fig.~\ref{fig:dig_examples}.

% \section{Comparison with Change Captioning}
% Change captioning methods~\cite{park2019robust,zhang2024differential} and DiG both take image pairs as input, but they address fundamentally different problems. Change captioning is a \emph{generative task}: it produces a natural language sentence describing what changed, supervised purely by text-level annotations. In contrast, DiG treats differential perception as a \emph{proxy task for foundational MLLM training}. Two distinctions are key. First, DiG supervises on \emph{dense spatial grounding} via bounding box sets, providing verifiable, fine-grained localization signals far beyond the coarse semantics of a caption. Second, DiG is not designed to excel at change captioning specifically; rather, the perceptual capability it cultivates generalizes to a broad range of downstream
% benchmarks unrelated to change description, something task-specific captioning models are not designed to achieve. In short, change captioning is an end task, while DiG is a perceptual pretraining framework that happens to leverage differential image pairs as its supervision signal.

\section{Analysis of Training Dynamics}

\begin{figure}[htbp]
  \centering
   \setlength{\abovecaptionskip}{0.cm}
   \setlength{\belowcaptionskip}{0.cm}
   \includegraphics[width=\linewidth]{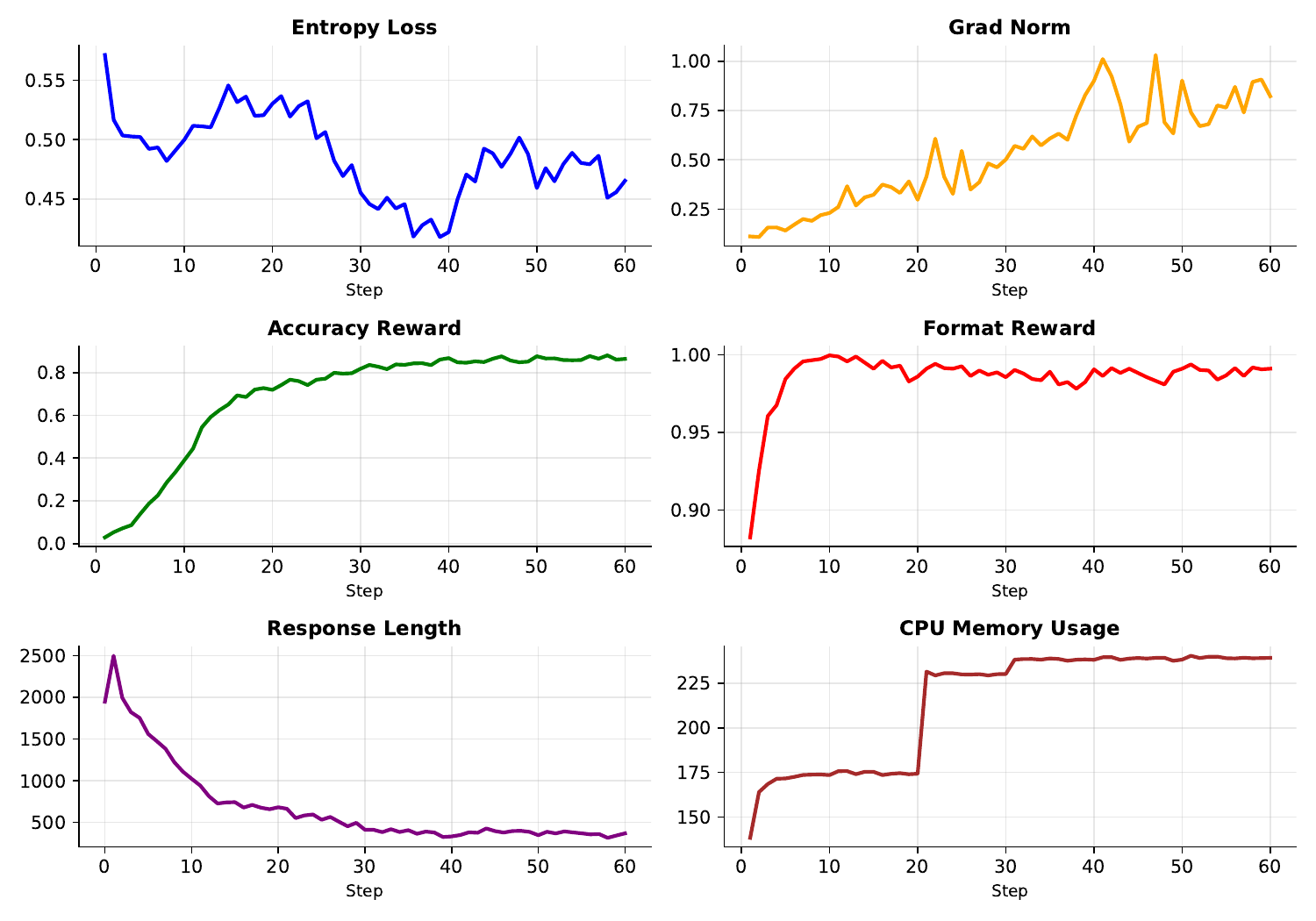}
   \caption{Training dynamics of the single-difference stage.}
   \label{fig:single_train}
\end{figure}

\begin{figure}[htbp]
  \centering
   \setlength{\abovecaptionskip}{0.cm}
   \setlength{\belowcaptionskip}{0.cm}
   \includegraphics[width=\linewidth]{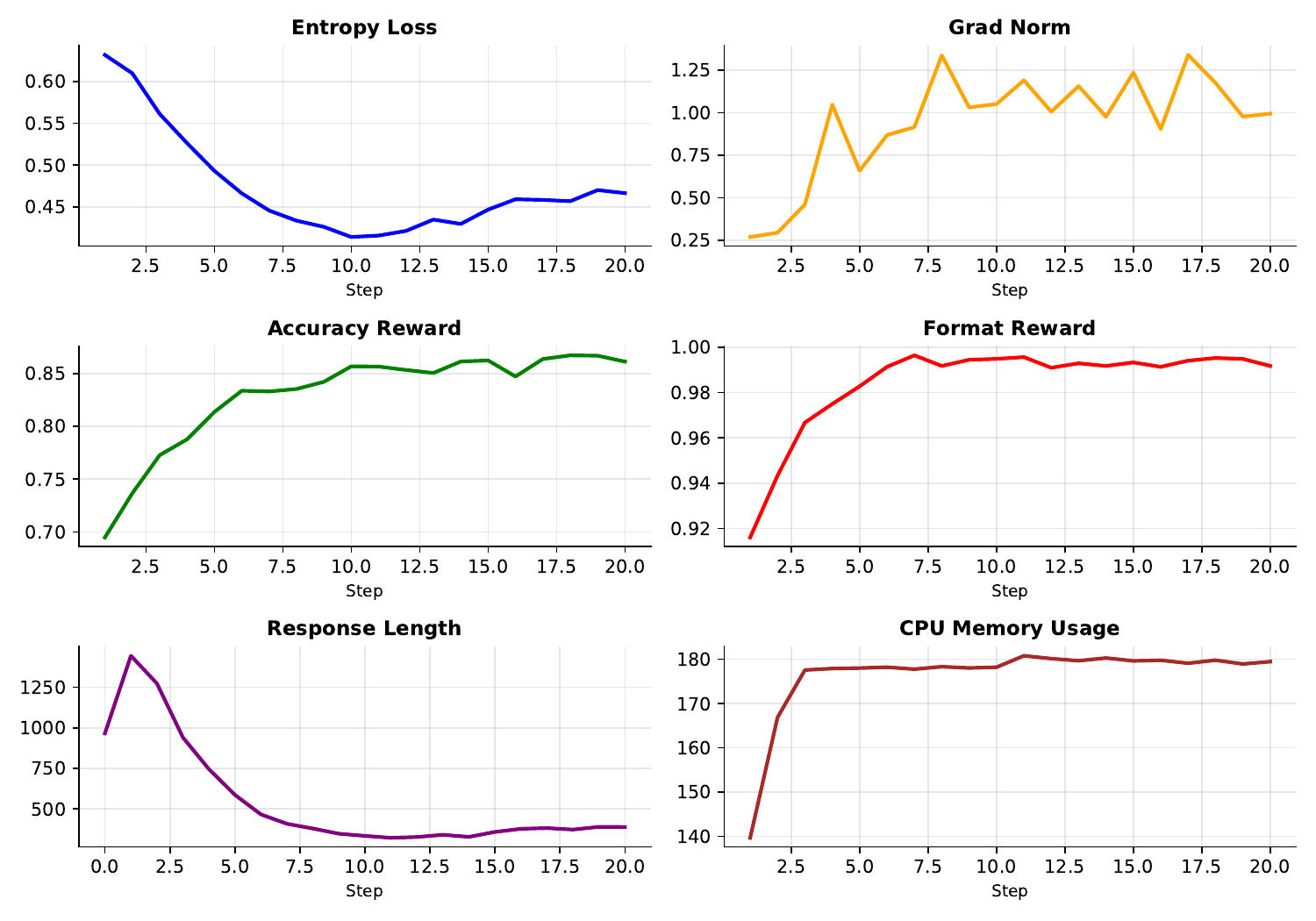}
   \caption{Training dynamics of the double-difference stage.}
   \label{fig:double_train}
\end{figure}

\begin{figure}[htbp]
  \centering
   \setlength{\abovecaptionskip}{0.cm}
   \setlength{\belowcaptionskip}{0.cm}
   \includegraphics[width=\linewidth]{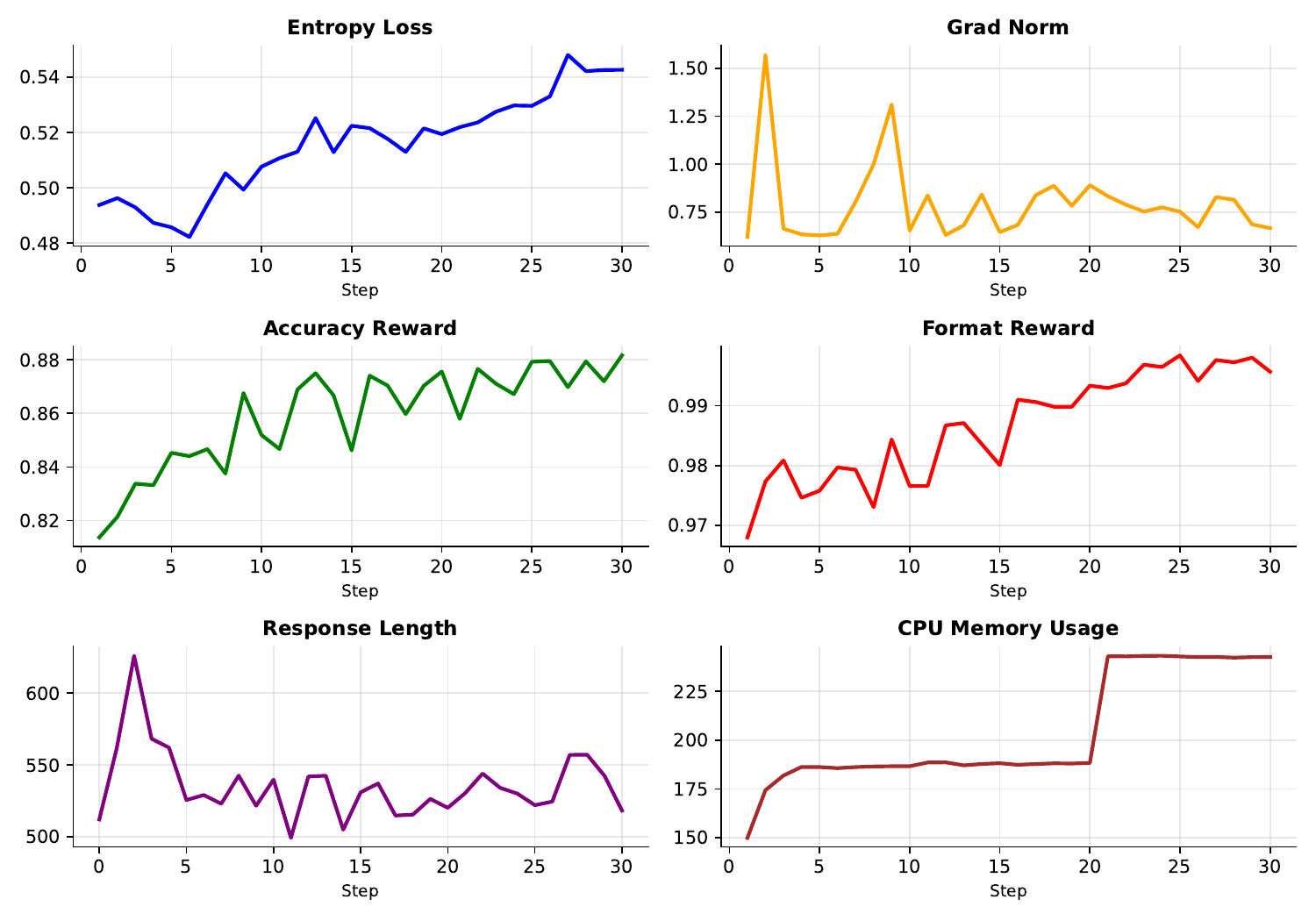}
   \caption{Training dynamics of the mix-difference stage. }
   \label{fig:mix_train}
\end{figure}

Figure~\ref{fig:single_train}, \ref{fig:double_train}, \ref{fig:mix_train}  illustrates the training dynamics of Qwen3-VL-8B-Thinking across the three curriculum stages. 
Overall, the entropy loss decreases steadily during the early stages and stabilizes as policy convergence occurs, indicating reduced exploration and more consistent generation behavior. 
The gradient norm remains bounded throughout training, confirming stable optimization.  
Both accuracy and format rewards exhibit monotonic improvement, with the latter quickly saturating near unity, suggesting the model rapidly adapts to the output structure constraints.  
Meanwhile, the average response length decreases as the model learns to produce more concise outputs, and CPU memory usage grows gradually with increased rollout complexity.  
Together, these trends demonstrate stable and efficient reinforcement post-training across stages.

\section{Case Study}

\begin{figure*}[t]
\centering
\begin{tcolorbox}[
    colback=white,           % 内容背景
    colframe=black!50,       % 边框浅灰
    colbacktitle=gray!25,    % 标题栏浅灰
    coltitle=black,          % 标题文字黑色
    fonttitle=\bfseries,     % 标题加粗
    title=Case Study of Differential Grounding (DiG),
    boxrule=0.4pt,
    arc=2pt,
    left=5pt, right=5pt, top=4pt, bottom=4pt,
    width=\textwidth         % 双栏宽度
]

\begin{minipage}{0.4\textwidth}
\centering
\begin{minipage}{0.9\textwidth}
\centering
\includegraphics[width=\linewidth]{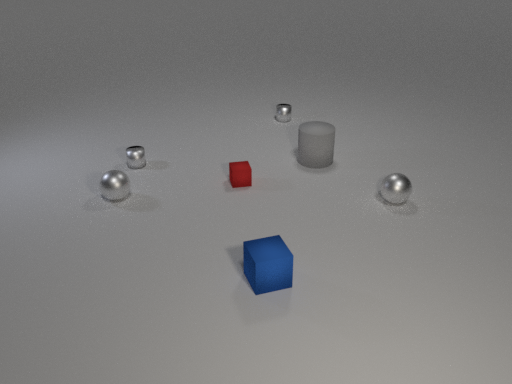}\\[-2pt]
{\small \textbf{Before}}\\[3pt]
\includegraphics[width=\linewidth]{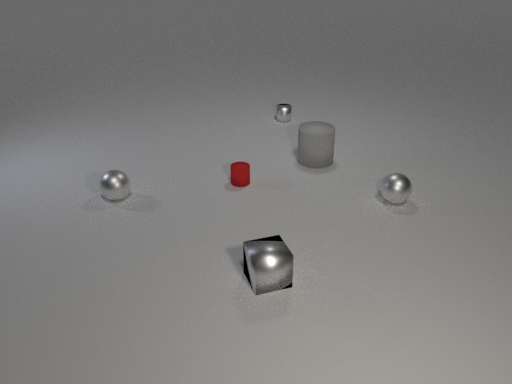}\\[-2pt]
{\small \textbf{After}}
\vspace{4pt}

% \caption*{\small \textbf{Input:} Example image pair from the DiG dataset used for difference localization.}
\end{minipage}

\vspace{4pt}
\caption*{\small \textbf{Input:} Example image pair from the DiG dataset. The model identifies and localizes visual differences between the two scenes.}
\end{minipage}
\hfill
\begin{minipage}{0.58\textwidth}
\small
\textbf{Model Output:}\\
Got it, let's compare the two images. First, the before image has a blue cube in the center, a red cube, and a small silver cylinder on the left. The after image has a metallic cube instead of the blue cube, and the red cube is now a red cylinder. Also, the small silver cylinder on the left is missing. Let's list the changes:

1. The red cube in the before image is now a red cylinder in the after. So the red cube's position is changed.
2. The blue cube in the before image is replaced by a metallic cube in the after. So the blue cube's position is changed.
3. The small silver cylinder (the one on the left) is missing in the after image. So that's a missing item.

Wait, need to check each object:

Before image objects:
- Silver sphere left
- Small silver cylinder (left)
- Red cube (center)
- Blue cube (bottom center)
- Small silver cylinder (top)
- Gray cylinder (right)
- Silver sphere right

After image objects:
- Silver sphere left (same)
- Red cylinder (where red cube was)
- Metallic cube (where blue cube was)
- Gray cylinder (same)
- Silver sphere right (same)
- Small silver cylinder (top) same
- Missing the small silver cylinder (left) and the blue cube.

So the changes are:
- The red cube (before) is now a red cylinder (after), so the red cube's position is a change.
- The blue cube (before) is replaced by a metallic cube (after), so that's a change.
- The small silver cylinder on the left (before) is missing in after, so that's a missing item.

Therefore, the regions to box are the positions of the red cube, blue cube, and the small silver cylinder on the left.
\texttt{</think>}

\texttt{<answer>}

[

    [446, 415, 490, 487],  // Red cube changed to red cylinder
    
    [476, 613, 573, 767],  // Blue cube replaced by metallic cube
    
    [238, 374, 289, 447]   // Small silver cylinder missing
    
]

\texttt{</answer>}

\end{minipage}

\end{tcolorbox}
\caption{\textbf{Case study of Differential Grounding (DiG).} 
Illustration of input image pairs and corresponding model outputs highlighting localized visual discrepancies.}
\label{fig:dig_case_study}
\end{figure*}

To further illustrate the behavior of our model, we conduct a qualitative case study on the Differential Grounding (DiG) task, as shown in Fig.~\ref{fig:dig_case_study}.  
Given a pair of images representing the ``before'' and ``after'' scenes, the model is required to identify and localize all visual differences between them.  
As demonstrated, the model accurately detects multiple types of changes, including object replacement and missing instances.  
Specifically, it recognizes that the red cube in the ``before'' image becomes a red cylinder, the blue cube is replaced by a metallic cube, and the small silver cylinder on the left disappears in the ``after'' scene.  
The predicted bounding boxes precisely correspond to these altered regions, confirming the model’s ability to jointly reason over visual semantics and spatial grounding.  
These results highlight the model’s fine-grained visual reasoning capacity and its robustness across complex difference configurations.

\end{document}